\pdfoutput =1
\documentclass[conference]{IEEEtran}
\IEEEoverridecommandlockouts
\usepackage{cite}
\usepackage{amsmath,amssymb,amsfonts}
\usepackage{algorithmic}
\usepackage{graphicx}
\usepackage{textcomp}
\usepackage{xcolor}
\usepackage{soul}
\usepackage{adjustbox}
\usepackage{booktabs}
\usepackage{tabularx}

\def\BibTeX{{\rm B\kern-.05em{\sc i\kern-.025em b}\kern-.08em
    T\kern-.1667em\lower.7ex\hbox{E}\kern-.125emX}}
\begin{document}

\title{ Visual Enhanced 3D Point Cloud Reconstruction from A Single Image}

\author{\IEEEauthorblockN{ Guiju Ping}
\IEEEauthorblockA{
\textit{Nanyang Technological University}\\
Singapore \\
Ping0008@e.ntu.edu.sg}
\and
\IEEEauthorblockN{Mahdi Abolfazli Esfahani}
\IEEEauthorblockA{
\textit{Nanyang Technological University}\\
Singapore \\
mahdi001@e.ntu.edu.sg}
\and
\IEEEauthorblockN{Han Wang }
\IEEEauthorblockA{
\textit{Nanyang Technological University}\\
Singapore \\
HW@ntu.edu.sg}
}

\maketitle

\begin{abstract}
Solving the challenging problem of 3D object reconstruction from a single image appropriately gives existing technologies the ability to perform with a single monocular camera rather than requiring depth sensors. In recent years, thanks to the development of deep learning, 3D reconstruction of a single image has demonstrated impressive progress. Existing researches use Chamfer distance as a loss function to guide the training of the neural network. However, the Chamfer loss will give equal weights to all points inside the 3D point clouds. It tends to sacrifice fine-grained and thin structures to avoid incurring a high loss, which will lead to visually unsatisfactory results. This paper proposes a framework that can recover a detailed three-dimensional point cloud from a single image by focusing more on boundaries (edge and corner points). Experimental results demonstrate that the proposed method outperforms existing techniques significantly, both qualitatively and quantitatively, and has fewer training parameters.

\end{abstract}

\section{Introduction}
3D reconstruction is an active research topic in the computer vision community and is a higher-level task than an image classification problem. Because it not only requires identifying objects but also recovers their full 3D shapes. In the past, 3D reconstruction was generally used in
industrial design, architecture design, anime film, medical modeling, robot navigation, robot object interaction, etc. With the emergence of machine learning and 3D vision technology, 3D reconstruction has been used in a wider range, like Augmented Reality (AR), Virtual Reality (VR), autonomous driving, remote sensing, mapping, 3D printing, and even online shopping. 3D reconstruction has crucial practical value and a promising future.
\begin{figure}[htb]
    \centering
    \includegraphics[width=\linewidth]{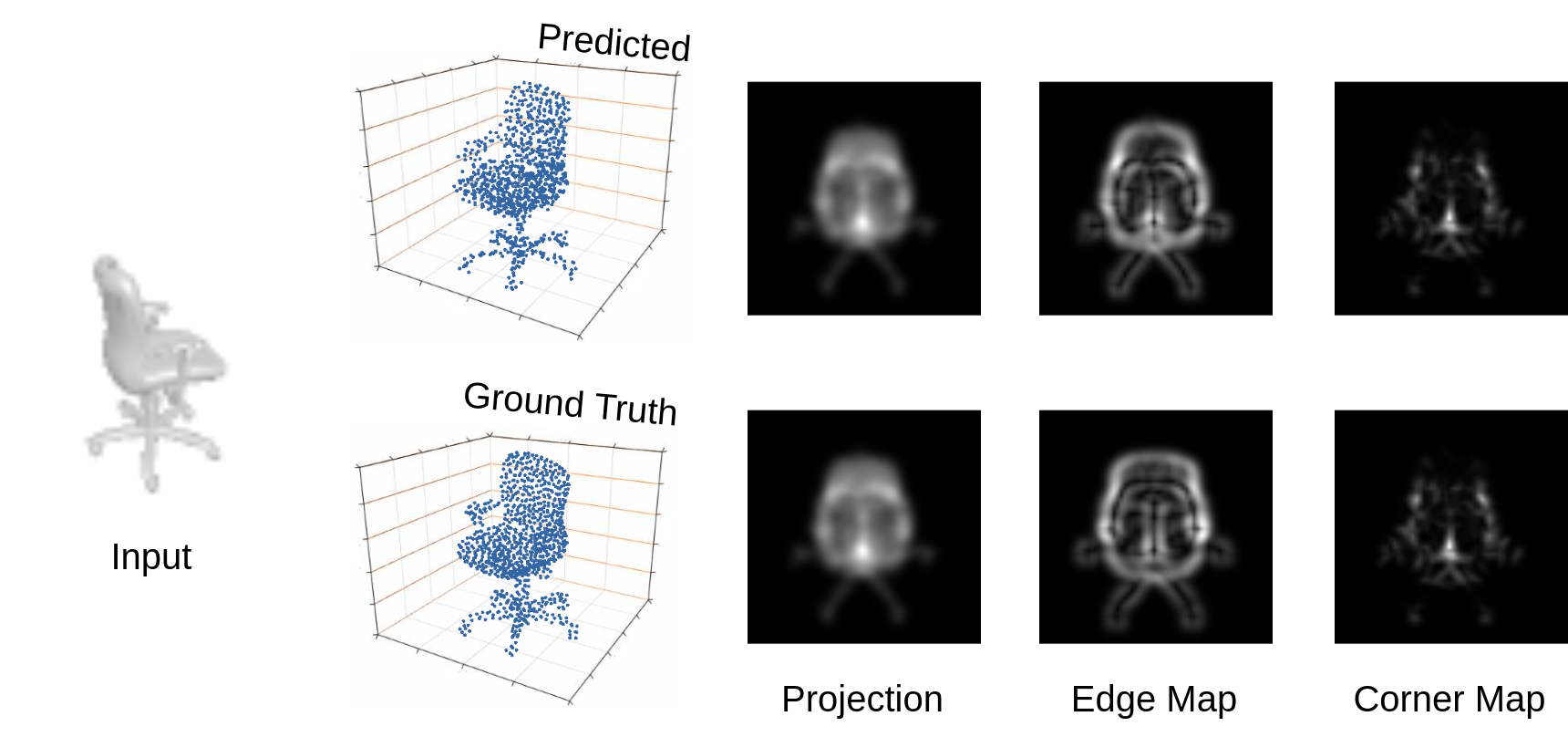}
    \caption{Our proposed method focuses more on the key points, such as edges and corners. It produces more visually satisfying reconstruction results. }
    \label{fig:my_label}
\end{figure}
However, capturing the 3D shape of an object is tedious and time-consuming. This usually requires specialized equipment and is operated by trained personnel. 
For instance, scanning real-life 3D objects from multiple angles with a 3D laser scanner,  or create manually with 3D modeling software, such as AutoCAD, Solidworks, etc. The process requires not only advanced hardware support but tedious optimization.

In contrast, 2D images are much easier and cheaper to obtain. Through the decades, researchers have developed many 3D reconstruction methods based on multi-view 2D  images \cite{kolev2008integration,kostrikov2014probabilistic,kar2017learning,wen2019pixel2mesh++}.  But in many circumstances, we can't get multiple paired images. It would be encouraging to recover 3D shapes from a single image which is a highly ill-posed and challenging question for machines.  Countless solutions would result in the same 2D projection, especially when others occlude the object or self occluded. However, we, as humans, could effortlessly decipher the underlying 3D structure from a single 2D image in natural conditions. How to make machines acquire 3D visual perception as humans?  

 Our research focuses on reconstructing 3D shapes using point cloud format. Although many 3D reconstruction frameworks based on point clouds have been proposed in the past few years,  we found that they usually use Chamfer Distance (CD) as the training loss, which gives equal weights to all points inside the point clouds. But for human visual perception, edges and corners are more important than flat points. They are the key factors that determine the structure of 3D objects. But annotation of such edge and corner points needs a lot of human resources and time. So in this paper, we proposed a framework that utilizes current advanced research on 2D images to strengthen the 3D reconstruction results. This is done by projecting the predicted point clouds onto 2D planes and use a Gaussian derivative and Harris corner detector to locate edge and corner points. All these processes must be differentiable in order to achieve end-to-end training. How to handle the discrete point cloud to get continuous projection is also a problem that needs to be handled carefully.

The rest of this paper is structured as follows: The related work is reviewed in Section \textrm{2}, and Section \textrm{3}  elaborates on our proposed methods. In Section \textrm{4}, extensive experiments are conducted, and the results are presented to verify the proposed methods. Finally, this paper is concluded in Section \textrm{5}.


\section{Related work}
Boosted by the rapid development of deep learning and the big data era,
a lot of single image 3D reconstruction methods have been proposed. Existing methods can be divided into four categories according to their output representations. 
\subsection{Volumetric }
The pioneer works, such as 3DR2N2 \cite{choy20163d}, Tl-emebedding \cite{girdhar2016learning}, and MarrNet \cite{wu2017marrnet},  are based on volumetric representations. They use voxel grids to represent the 3D shape, as voxel grids are intuitive, regular, and naturally suitable for neural network training. 3DR2N2 \cite{choy20163d} use a Recurrent Neural Network (RNN) to fuse multiple feature maps extracted from input images sequentially and gradually recover the 3D shapes. Their framework can be used in both single image reconstruction and multiple-views reconstruction. However, if the input order changes, RNN-based approaches are unable to produce consistent reconstruction results. Tl-embedding \cite{girdhar2016learning} first trained a 3D autoencoder using cross-entropy loss. Then they trained an image encoder with AlexNet to map the latent space of the 3D auto-encoder. MarrNet \cite{wu2017marrnet} used a depth map as a bridge to reconstruct 3D shapes. All these volumetric-based methods suffer from a major drawback, the high memory footprint. The memory grows cubically as the resolution goes higher,  which limited its capability to get higher reconstruction quality. Normally used resolutions are $32^3$ or $64^3$, which are too low to retain detail structures.

\subsection{Point Cloud}
Gradually researchers started to reconstructed 3D shapes in a point cloud format. Point cloud representations are much more memory efficient as they only store the information on the object surfaces. Pint cloud is unordered data.  Researchers proposed  PointNet \cite{fan2017point} and PointNet++ \cite{qi2017pointnet++} to address this problem. 
3D-LMNet \cite{mandikal20183d} use the PointNet structure to build an auto-encoder to obtain the latent representation of 3D point clouds. Then, they use another network to match the latent representation of 2D images and paired 3D point clouds. 
PSGN \cite{fan2017point} introduced the Hourglass convolution network structure to obtain a more vital presentation. Their design can make better use of global and local information by utilizing several parallel predictive branches.

Such a complex model is highly flexible and excels in describing complex structures, but it also inevitably consumes a lot of memory.  CAPNet \cite{navaneet2019capnet} proposed a framework that could recover 3D point cloud with only 2D masks as supervision. This is done by projecting the predicted point clouds onto 2D planes and use their proposed Affine loss as training loss. GAL \cite{jiang2018gal} designed a hybrid loss that combines Chamfer loss, geometric loss, and conditional adversarial loss together. The geometric loss is computed on the 2D projected images from different views and different resolutions.    
Other researches \cite{jiang2018gal,kurenkov2018deformnet,mandikal2019dense,sun2019ssl,lu2019attention} 
have also achieved varying degrees of success. However, all these researches treat every point inside the point cloud equally. They didn't emphasize the key points. While they could give good evaluation metrics, their reconstruction results look visually unsatisfactory.

\subsection{Mesh}
 Compared with the point cloud, the advantage of mesh representation is that it can preserve connectivity. Mesh-based single image 3D reconstruction always works in a deform and regeneration way. Pixel2Mesh \cite{wang2018pixel2mesh} use an ellipsoid mesh as a starting point and learns the deform for each vertex point. Their model contains three mesh deformation blocks. Each deformation block increases mesh resolution and estimates vertex locations.  However, mesh-based 3D reconstruction cannot handle objects with complex topological changes. This limited their generalizability. Mesh-based single image reconstruction normally works in a single category domain \cite{kar2015category,kanazawa2018learning}. Recently, Tang et al. proposed SkeletonNet \cite{tang2019skeleton,tang2020skeletonnet}, which uses object skeleton as a bridge to preserve topology. It can produce good reconstruction results, but it requires a complex pre-processing stage to obtain 3D skeletons. Point cloud representation has more freedom compared to mesh representation.

\subsection{Implicit Surface}

The implicit surface is a method that has been used to represent 3D shapes only in recent two years \cite{xu2019disn,mescheder2019occupancy,chen2019learning,genova2019learning,yamashita20193d,thai20203d}, and is first proposed by Park et al. \cite{park2019deepsdf}. Implicit surface based approaches can produce a more accurate result with arbitrary resolutions. But they require post-processing algorithms such as marching cube \cite{lorensen1987marching} to extract underlying structures. Marching cube is an expensive and time-consuming operation. Previous methods ( volumetric, mesh, point cloud) only require to pass the neural network once, while implicit surface based methods need each sampled point to go through the neural network; It highly limited their ability to be applied in real-world applications.

\section{Proposed Method}
\begin{figure*}[htb]
    \centering
    \includegraphics[width=\linewidth]{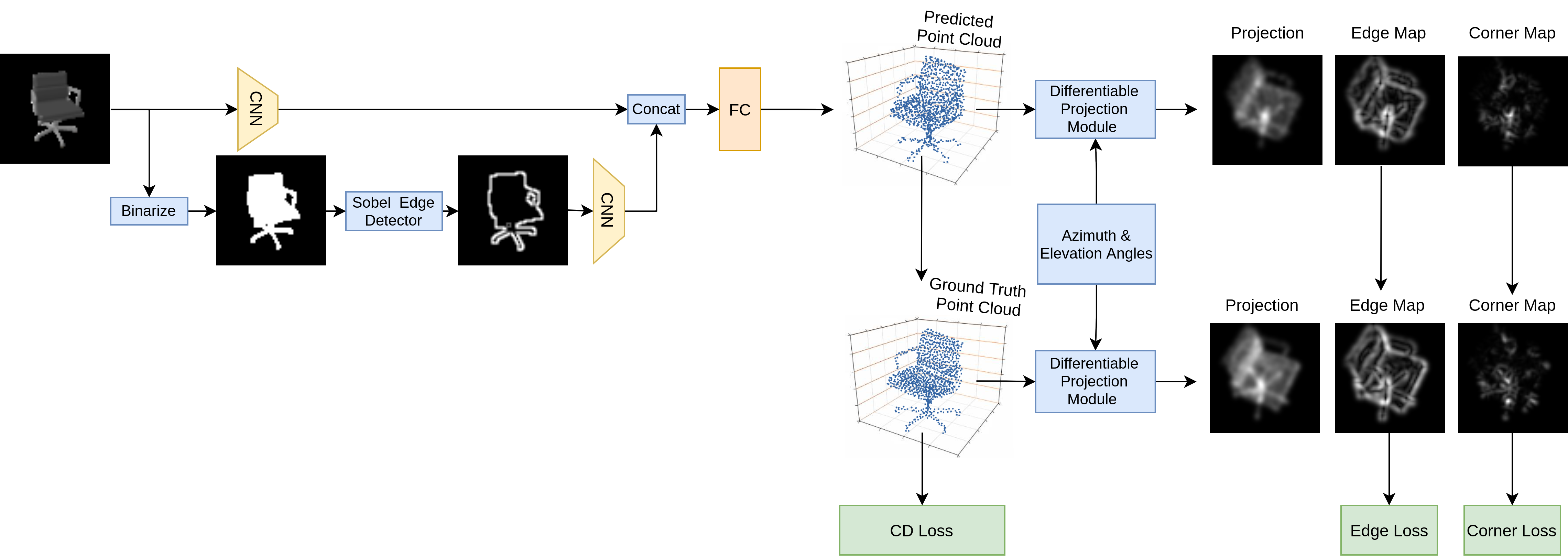}
    \caption{The pipeline of 3D-VENet. The network first binarizes the input image and finds its edges through a Sobel edge detector, then two branches of CNN are applied to extract the features in the input image and its edges. After that, the two brunches' features are concatenated and pass to fully connected layers to predict the final 3D point clouds. Then the differentiable projection module is applied, and losses are calculated to guide the training process. The network can be trained end to end.  }
    \label{fig:pipeline}
\end{figure*}
\subsection{Problem definition}
Given a single image of an object, $I$, this paper aims to reconstruct its 3D point cloud representation $\hat{P}$, via the deep learning framework. Hence, this paper investigates the neural network $f(.)$ to reconstruct the point cloud from a single image,  $\hat{P}=f(I)$ .

\subsection{Overview}

 CNN is utilized in the first step to extract rich features from 2D images. We believe the edges inside the input image are essential for the network to recover the full 3D shape. So we employ a two-branch structure. One branch takes the original image as input, and another branch takes the edge map as input. The proposed system does not introduce an additional workload for dataset collection or annotation because the edge map is automatically acquired by image processing during the training. Then, the feature representations from these two branches are concatenated. After that, fully connected layers are applied to reconstruct the 3D point cloud from extracted features. Once the predicted 3D point cloud is obtained, a projection module is used to get the 2D projected images. This projection module is differentiable, thus allows backpropagation. Unlike previous researches \cite{navaneet2019capnet,jiang2018gal,sun2019ssl}, we didn't compute any loss directly on the projected images. It is because computing losses on the projected images will still backpropagate to every point inside the 3D point cloud like what CD loss does.  We want to add more emphasize on edge/corner points. So instead of calculating loss directly on projected images, we compute losses on edge/corner maps.
  The Overall framework is shown in Figure \ref{fig:pipeline}.

\subsubsection{Network Design}\label{sec:network_design}
For 2D image feature extraction, we use convolution neural networks to map the input images to a 512-dimensional latent vector. Each convolution layer is activated with the ReLU activation function and normalized using the L2 norm.  After feature extraction, we use three fully connected layers of size [1024,1024,$N \times 3$] to predict the reconstructed point clouds. $N$ is the total number of points inside a point cloud, which is set to 1024 in all of our experiments. Details of feature extraction and decoder structure are shown in supplementary material.

\subsubsection{Differentiable Projection Module}

The projection follows a classical pinhole camera projection model defined by the intrinsic parameters matrix $K$, rotation $R \in S O(3)$ and translation $t \in \mathbb{R}^{3}$ between world and camera coordinate systems:
\begin{equation}
\widetilde{P}=K[R \quad t] P
\end{equation}
where $P$ is a given 3D point cloud. 
$\widetilde{P}$ is a matrix that stores the projected  location of each point.

\begin{figure*}[htb]
    \centering
    \includegraphics[width=0.9\linewidth]{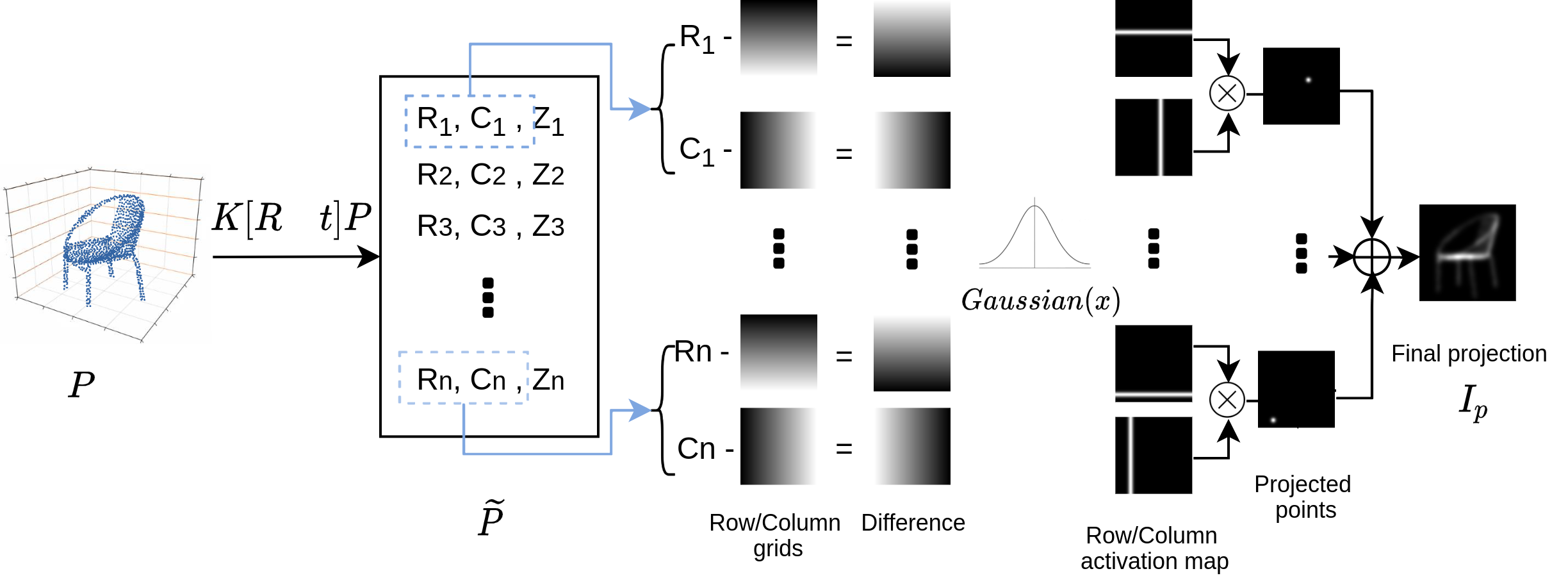}
    \caption{Differentiable Projection module. $P$ is the given 3D point cloud. $\widetilde{P}$ is a matrix that stores the projected  location of each point. $R_{1}$ and $C_{1}$ are the projected row index and column index of the first point inside the 3D point clouds. $n$ is the total points number inside the point clouds, which is set to 1024 in all our experiments.  (All the images shown in the above figure are normalized to [0,255] for better visualization.)
 }
    \label{fig:projection}
\end{figure*}

 For all of our experiments, we set up a virtual camera 2.5 unit away from the obejcts along z axis. 
 so $t$ is defined as 
 \begin{equation}
   t=\left[ 0,0,2.5\right]  
 \end{equation}
 The intrisic parametrs matrix $K$ for the virtual camera is defined as
\begin{equation}
K=\left[\begin{array}{ccc}
120 & 0 & -32 \\
0 & 120 & -32 \\
0 & 0 & 1
\end{array}\right]
\end{equation}
This $K$ matrix and $t$ will ensure the projected points fall into a 64 $\times$ 64 image.
After getting the $\widetilde{P}$, the traditional way is to use a scatter function to obtain the projected images. But scatter function is not differentiable, thus doesn't allow backprogatation. 
Inspired from CAPNet \cite{navaneet2019capnet}, we designed a continuous approximation projection module. 
To better capture detailed information and retain more information on the projected image, we removed the $tanh$ layer from their design. This is also beneficial for backpropagation; since the $tanh$ layer has a gradient of $0$ when saturated. We first build two grids; row grid and column grid. Both of them have the same size as the original images, which is 64 $\times$ 64 in our settings. The row grid is filled with 0 to 63 from row 0 to row 63. The column grid is filled with 0 to 63 from column 0 to 63. Then each point, $\widetilde{P}_{i}$, inside  $\widetilde{P}$, we find the difference between  $\widetilde{P}_{i}$ and the row/column grids. After that,  Gaussian function (Equation.\ref{eqn:gaussian})
\begin{equation}
Gaussian(x)=e^{-\frac{x^{2}}{2\sigma^{2}}}
\label{eqn:gaussian}
\end{equation}
is applied to obtain the row/column activation maps. Then the row and column maps are multiplied to get the final projection points. Finally, the projection images of all the 1024 points are summed together to obtain the final projected images $I_p$. Details are illustrated in Figure \ref{fig:projection}.

\begin{figure}[htb]
    \centering
    \includegraphics[width=\linewidth]{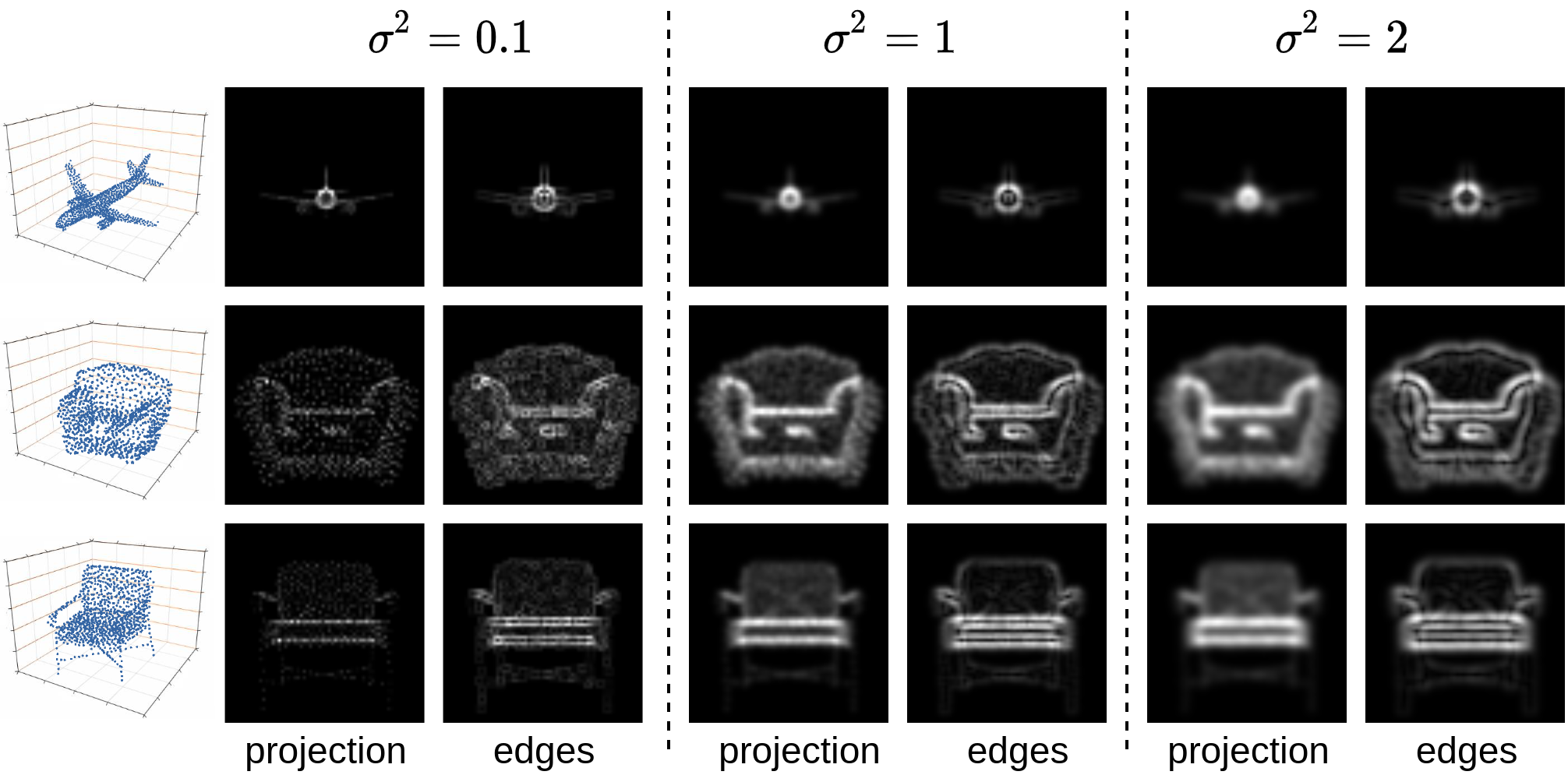}
    \caption{Projection with different $\sigma^{2}$ values. Column 1 is the 3D point clouds. Column 2 to column 7 are the projection results and edges when using different $\sigma^{2}$ values. (All images are normalized to [0,255]).
    }
    \label{fig:compare_sigma}
\end{figure}

 For the Gaussian activate function (Equation \ref{eqn:gaussian}), how to set its  $\sigma^2$ value is worth considering. Different from Capnet \cite{navaneet2019capnet}, we used a much bigger $\sigma^2$, for our Gaussian function (Equation \ref{eqn:gaussian}). This is to ensure the projected images look connected even when the point cloud is very sparse.
 As shown in Figure \ref{fig:compare_sigma}, when $\sigma^2=0.1$ the shape details can be retained, but the neighboring points cannot be connected in the final projected images. As we all know, point clouds are sampled values from the underlying 3D surface. We can sample infinite point clouds from one surface. If we directly apply loss on these projected images, it will incur some losses even if the ground truth point cloud and predicted point cloud have the same underlying surface. So we choose to use a bigger $\sigma^2$ value to eliminate the incurring of this unnecessary loss to let the network focus more on the underlying structures. Bigger $\sigma^2$ value also helps to get a clear edge map, which is essential for computing our $Edge Loss$ as elaborated in Section \ref{sec:loss}.

\subsection{Construct Edge Maps and Corner Maps}\label{sec:edges}

We construct two $3 \times 3$ Gaussian derivative kernels and convolve these Gaussian derivative kernels with the projected images, $I_p$, to find edge maps as follow:

\begin{equation}
\begin{array}{l}
I_{x}=I_p * \frac{\partial}{\partial x}(G) \\
I_{y}=I_p * \frac{\partial}{\partial y}(G)
\end{array}
\end{equation}
where $G$ is a 2D Gaussian function.

The edge map is computed using

\begin{equation}
    Edge Map=|{I_{x}}|+|{I_{y}}|
\end{equation}
Here, we didn't use square root to find edge maps. It's because we found that the square root function has a large gradient when input values are near zeros. This would lead to gradient exploration during taring.
After getting $I_{x}$ and $I_{y}$, we use the Harris Corner Detector algorithm to locate corners. It can be defined mathematically as


\begin{equation}
\begin{aligned}
M &=\sum_{(x, y) \in W}\left[\begin{array}{cc}
I_{x}^{2} & I_{x} I_{y} \\
I_{x} I_{y} & I_{y}^{2}
\end{array}\right] \\
&=\left[\begin{array}{cc}
\sum_{(x, y) \in W} I_{x}^{2} & \sum_{(x, y) \in W} I_{x} I_{y} \\
\sum_{(x, y) \in W}I_{x} I_{y} & \sum_{(x, y) \in W} I_{y}^{2}
\end{array}\right]
\end{aligned}
\end{equation}

\begin{equation}
\operatorname{trace}(M)=m_{11}+m_{22}
\end{equation}
\begin{equation}
Corner Map=\frac{\operatorname{det}(M)}{\operatorname{trace}(M)+eps}
\end{equation}
where $W$ is a 3 by 3 Gaussian kernel. A small $eps$ value is added to prevent runtime zero-division errors.

\begin{figure}[htb!]
    \centering
    \includegraphics[width=0.7\linewidth]{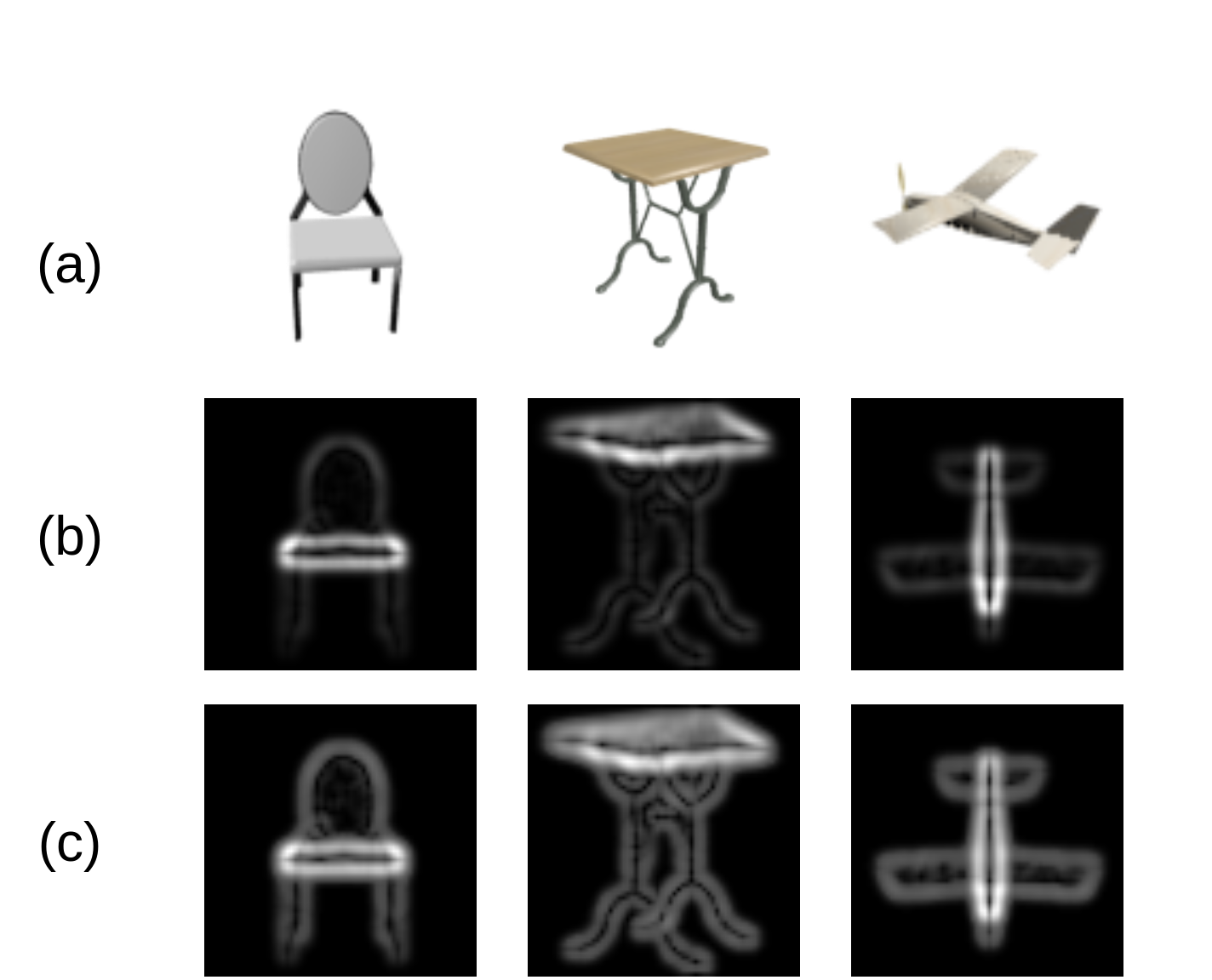}
    \caption{Edge maps before suppression (b) and after suppression (c). All images are normalized to [0,255] for better visualization.} 
    \label{fig:supression}
\end{figure}

After obtaining the edge and corner maps, we found these maps have a very high response at the locations where the points' density is high, such as the chair seats and table surfaces ( as shown in Figure \ref{fig:supression}). Therefore, we apply a suppression function to re-map the high value to a lower range. What we do is we first normalize edge and corner maps to [0,1], then for each pixel value $x$ in edge and corner maps, we apply the following function

\begin{equation}
Suppression(x) =\left\{\begin{array}{ll}
x, & \text { if } x \leq0.1\\
0.1+0.3(x-0.1), & \text { if } x>0.1
\end{array}\right.
\end{equation}
\subsection{Loss Function}\label{sec:loss}
As point cloud is an unordered data format, using the regular $\mathrm{L} 1$ or $\mathrm{L} 2$ loss can cause regression difficulty. Chamfer Distance (CD) is a commonly used metric for comparing two point clouds' similarity.
We define the CD between the predicted point cloud $\hat{P}$ and ground truth point cloud $P$ as:
\begin{equation}
\label{eqn:cd}
\begin{aligned}
&CD\left(\hat{P}, P\right)=\sum_{x \in \hat{P}} \min _{y \in P}\|x-y\|_{2}^{2}+\sum_{y \in P} \min _{x \in \hat{P}}\|x-y\|_{2}^{2}
\end{aligned}
\end{equation}

For each point, the CD's algorithm finds the nearest neighbor in the other set and sums the squared distances up. CD is continuous and piecewise smooth. The range search for each point is independent, thus trivially parallelizable.  CD produces reasonable high-quality results in practice.  But it gives equal weights to all points in the 3D point cloud, but for human visual perception, edges and corners are more critical. 

Since our projection module uses a big $\sigma^2$ to blur the discretized points to get a connected 2D projection, the second issue of Chamfer loss can be solved. A blurred and smoothed image also helps to get a clear edge map. To cater to the first drawback of Chamfer Distance, we designed $Edge Loss$. Let $E$ be the ground truth projected edges, and $\hat{E}$ be the predicted edges, then the Edge loss can be defined as
\begin{equation}
    Edge Loss= L_1(E-\hat{E})
\end{equation}
Similarly, Let $C$ be the ground truth projected corner maps, and $\hat{C}$ be the predicted corners, $CornerLoss$ is defined as
\begin{equation}
    Corner Loss= L_1(C-\hat{C})
\end{equation}
Our final loss is defined as:
\begin{equation}
    Loss=CD+\lambda_1 *EdgeLoss+\lambda_2 *Corner Loss
\end{equation}
where $\lambda_1$, and $\lambda_2$ are tunable parameters. In this paper we have considered $\lambda_1 =20$ and $\lambda_2=10$.

\section{Experiments and Results}
\subsection{Dataset}
Experiments based on the following two well-known datasets are applied in this section to evaluate the proposed 3D-VENet. 
\begin{itemize}
    \item \textbf{1. ShapeNet} \cite{chang2015shapenet} is used for the training and evaluation of the proposed method. We adapted the same train and test split as in 3D-LMNet\cite{mandikal20183d}. There are a total of 840528 images for the training set, which are rendered from 35022 3D models, 24 images for each model. The test set of ShapeNet\cite{chang2015shapenet} dataset contains 210288 images, which are rendered from 8762 3D models. The models come from 13 categories; The data distribution of data can be seen in the supplementary material.
    \item  \textbf{2. Pix3D} \cite{sun2018pix3d} dataset is just used for testing. It contains real-world images from 9 categories. Among the nine categories, three categories (chair, sofa, table) are co-occurred with our training data. So we test our trained model on these three categories. We filter out the occluded images as Mandikal et al. \cite{mandikal20183d} to have a fair comparison. Eventually, there are a total of 2892 chairs, 1092 sofas, and 738 tables.

\end{itemize}

\subsection{Implementation details}
All input images for our network are $64 \times 64$, and all 3D point clouds are normalized within a bounding box of unit 1 before calculating the losses. The learning rate is set to 0.00005, and an Adam optimizer is utilized. The mini-batch size is set to 12. 

\begin{figure}[htb!]
    \centering
    \includegraphics[width=\linewidth]{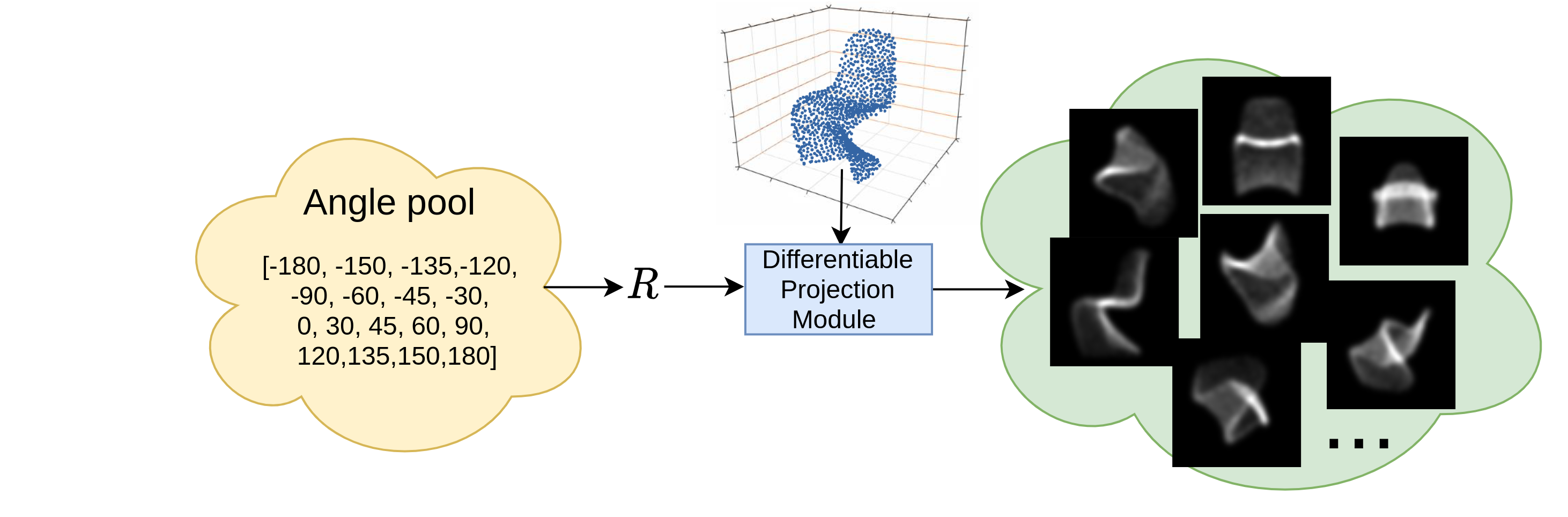}
    \caption{Angle pool and resulted projections.}
    \label{fig:four_angles}
\end{figure}

 For every batch, we randomly select four groups of azimuth and elevation angles from our angle pool to construct four rotation matrices, $R$, to obtain projections from four angles, as shown in Figure \ref{fig:four_angles}. Both the ground truth projection and predicted projection are obtained during training. After the model saturated, we changed $\lambda_1$ to $2$, and $\lambda_2$ to $0.2$, and the model is trained for another epoch. This is to correct some isolated outliers. Because such outliers will result in fragile edges, which can only gain small weights during backpropagation. Reducing $\lambda_1$ and $\lambda_2$ will let CD loss takes more effect. We also tried to set $\lambda_1=\lambda_2=0$, but we notice this will increase EMD loss. So we maintain a low $\lambda_1$ and $\lambda_2$ to correct the outliers.
\begin{figure*}[htb]
    \centering
    \includegraphics[width=\linewidth]{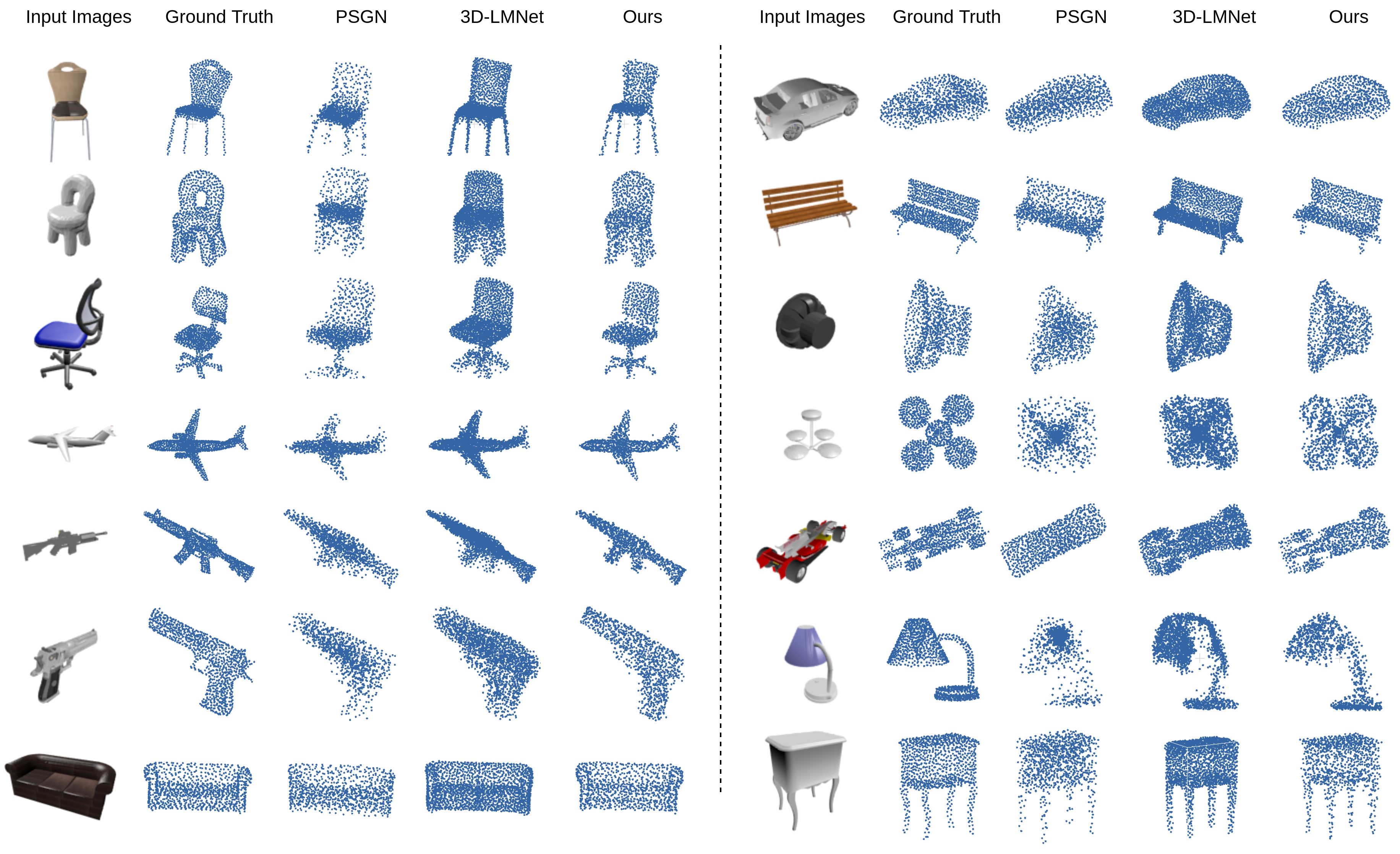}
    \caption{Single image 3D reconstruction test results on ShapeNet dataset.3D-Lmnet are viewed with 2048 points, the rest are viewed with 1024 points.}
    \label{fig:shapenet}
\end{figure*}

\begin{table*}[htb]

\centering

\setlength{\tabcolsep}{5mm}{
\begin{tabularx}{0.8\textwidth}{@{}l|lll|lll@{}}
\hline
& \multicolumn{3}{c|}{Chamfer Distance} & \multicolumn{3}{c}{EMD}     \\ \hline
Category       & PSGN\cite{fan2017point}     & 3D-LMNet\cite{mandikal20183d}   & Ours     & PSGN\cite{fan2017point}   & 3D-LMNet\cite{mandikal20183d} & Ours \\ \hline \hline
airplane  & 3.74 &3.34 &\textbf{3.09} & 6.38 & 4.77&\textbf{3.56} \\
bench & 4.63 & 4.55 & \textbf{4.26}& 5.88 & 4.99&\textbf{4.09} \\
cabinet & 6.98 & 6.09 &\textbf{5.49}  & 6.04 & 6.35&\textbf{4.69} \\
car & 5.20 & 4.55 &\textbf{4.30}  & 4.87 & 4.10&\textbf{3.57} \\
chair & 6.39 & 6.41 &\textbf{5.76}  & 9.63 &8.02 &\textbf{6.11}\\
lamp  & 6.33 & 7.10 & \textbf{6.07} & 16.17 & 15.80&\textbf{9.97} \\
monitor  & 6.15 & 6.40 &\textbf{5.76 }  & 7.59 & 7.13&\textbf{5.63} \\
rifle & 2.91 & 2.75 &\textbf{2.67} & 8.48 & 6.08 &\textbf{4.06}\\
sofa  & 6.98 & 5.85 &\textbf{5.34} & 7.42 & 5.65 &\textbf{4.80}\\
speaker  & 8.75 & 8.10 &\textbf{7.28}  & 8.70 & 9.15&\textbf{6.78} \\
table  & 6.00 & 6.05 &\textbf{5.46}
  & 8.40 & 7.82 &\textbf{6.10}\\
telephone  & 4.56 & 4.63 & \textbf{4.20}
  & 5.07 & 5.43 &\textbf{3.61} \\
vessel  & 4.38 & 4.37 & \textbf{4.22}  & 6.18 & 5.68&\textbf{4.59} \\
\hline mean  & 5.62 & 5.40 &\textbf{4.92}

 & 7.75 & 7.00&\textbf{5.20}
 \\
 \hline
\end{tabularx}}

\caption{Single view reconstruction results on ShapeNet\cite{chang2015shapenet}. The metrics are calculated on 1024 points after pefroming ICP alignment. All metrics are scaled by 100.}

\label{tab:shapenet}
\end{table*}

\begin{figure*}[htb]
    \centering
    \includegraphics[width=\linewidth]{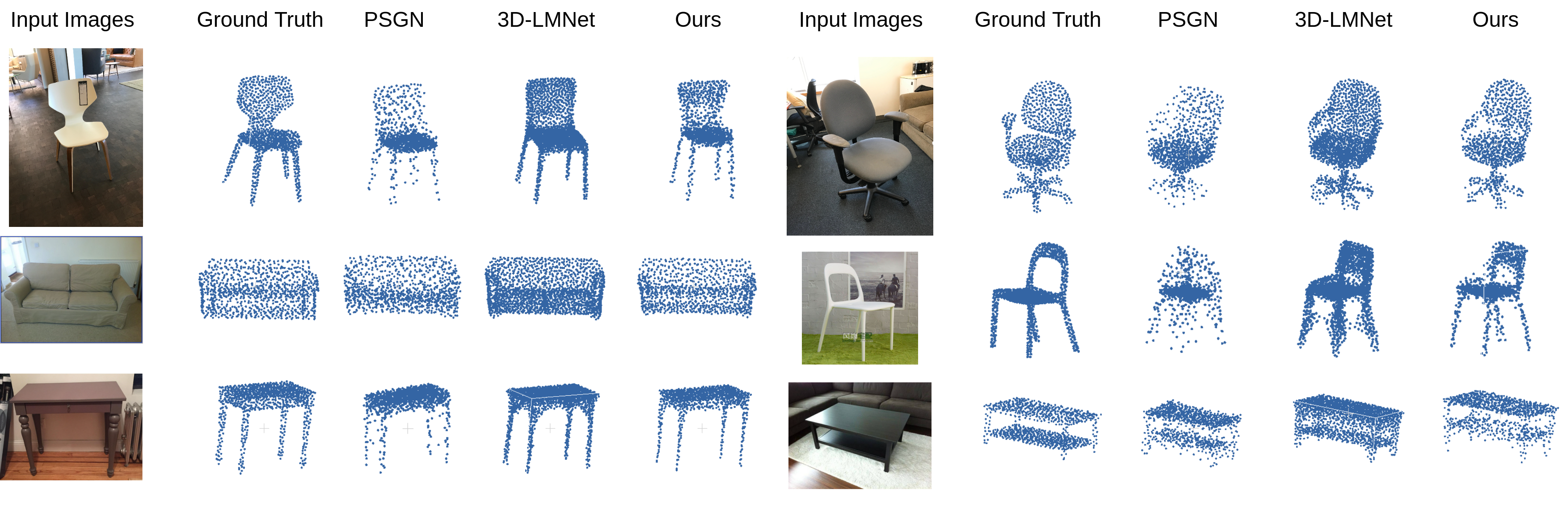}
    \caption{Single image 3D reconstruction results on Pix3D dataset.}
    \label{fig:pix3d}
\end{figure*}

\subsection{Evaluation Methodology}
We report the testing results using two metrics, CD (Equation \ref{eqn:cd}) and EMD. EMD between the ground truth point cloud $P$ and predicted point cloud $\hat{P}$ is defined as:

\begin{equation}
E M D\left(P, \hat{P}\right)=\min _{\phi: P \rightarrow \hat{P}} \sum_{x \in P}\|x-\phi(x)\|_{2}
\end{equation}
where $\phi: P \rightarrow \hat{P}$ is a bijection.

All point clouds are normalized to a unit cube. The predicted point cloud is aligned with the ground truth point cloud using ICP \cite{besl1992method} algorithm before calculating the metrics. For both CD and EMD metrics, the smaller value is better.

\subsection{Results}

 We compared our method with two well-known state-of-the-art algorithms, PSGN \cite{fan2017point} and 3D-LMNet\cite{mandikal20183d}.  Both of them are trained with the same dataset and train/test split. Since 3D-LMNet predicts 2048 points,  we randomly sampled 1024 points for calculating the metrics. The quantitative results are shown in Table \ref{tab:shapenet}. 
Please note that the input image size is $128 \times 128$ for both  PSGN \cite{fan2017point} and  3D-LMNet \cite{mandikal20183d}, while our input image size is $64 \times 64$. But even with smaller input size, our model  outperform 3D-LMNet \cite{mandikal20183d} and PSGN  \cite{fan2017point} in both CD and EMD metrics. Our model perform better in all 13 categories, especially in EMD metric. A lower EMD score also correlates with better visual quality and encourage points to lie closer to the surface. Our trainable parameters are $81.7\%$ of 3D-LMNet \cite{mandikal20183d} and $43.3\%$ of PSGN \cite{fan2017point}; which demonstrates a faster and more robust solution.

 The qualitative results are shown in Figure \ref{fig:shapenet}. From the Figure, we can see that the proposed model captures the input images' structure better, especially the curved contours of the underlying 3D objects. This shows that the proposed model pays attention to the edges/corners of the objects. More visual results are available in the supplementary material. However, the proposed model still cannot recover tiny holes because such small features are blurred by Gaussian function during projection. If ground truth projected images be available with clear holes, the results could be improved.The visual results for Pix3D are shown in Figure \ref{fig:pix3d}

\section{Conclusion}
This work proposed a framework to reconstruct a 3D point cloud from a single image. By applying a differentiable projection module, edge/corner points are located inside the projected images. The experiments show that the proposed framework can focus on edges/corners and eventually get more visually satisfying results. The proposed visual-enhanced method can also be used in a 3D generative model to generate high-quality point clouds. It can also be used in self-supervised 3D reconstruction if the input images' viewpoint is known.

{\small
\bibliographystyle{unsrt}
\bibliography{un}
}
\end{document}